\documentclass[runningheads]{llncs}
\usepackage{amsmath}
\usepackage{graphicx}
\usepackage[caption=false]{subfig}
\usepackage{algorithmic}
\begin{document}
%
\title{A Framework for Imbalanced Time-series Forecasting}
%
%
\author{Luis P. Silvestrin\inst{1}\orcidID{0000-0002-5759-1986} \and
Leonardos Pantiskas\inst{1}\orcidID{0000-0002-4898-5334} \and
Mark Hoogendoorn\inst{1}\orcidID{0000-0003-3356-3574}}
\authorrunning{L.P. Silvestrin et al.}
%
\institute{Computer Science Department, Vrije Universiteit Amsterdam, NL}
\maketitle              
\begin{abstract}
Time-series forecasting plays an important role in many domains. 
Boosted by the advances in Deep Learning algorithms, it has for instance been used to predict wind power for eolic energy production, stock market fluctuations, or motor overheating. 
In some of these tasks, we are interested in predicting accurately some particular moments which often are underrepresented in the dataset, resulting in a problem known as \emph{imbalanced regression}. 
In the literature, while recognized as a challenging problem, limited attention has been devoted on how to handle the problem in a practical setting. 
In this paper, we put forward a general approach to analyze time-series forecasting problems focusing on those underrepresented moments to reduce imbalances. 
Our approach has been developed based on a case study in a large industrial company, which we use to exemplify the approach.

\keywords{Imbalanced regression \and Deep Learning \and Time-series forecasting \and Multivariate time-series.}
\end{abstract}
\section{Introduction}
Due to the recent advances in artificial intelligence research, 
the task of time-series forecasting is being increasingly tackled with machine learning and deep learning techniques. 
There has been a large number of approaches suggested, ranging from relatively simple machine learning models\cite{ahmed2010empirical} to a variety of deep learning models\cite{torres2021dlsurvey}. 
Those approaches have been utilized in a broad spectrum of forecasting tasks, such as wind power forecasting, stock market prediction and motor temperature prediction\cite{torres2021dlsurvey}.
In the above examples of tasks, as well as in multiple other applied cases, some samples are more crucial from the point of view of the user and thus would require a more accurate prediction from a model compared to its average performance. 
At the same time, those data points may be scarce in the training data. Hence, if left unattended performance might be worse than average for that data, which is highly undesirable. This issue is characterized as imbalanced regression, and so far has been addressed with data pre-processing or ensemble model methods\cite{brancoSurveyPredictiveModeling2016}.

Despite the existing methods to tackle imbalanced regression problems, it is still a non-trivial task to data scientists and machine learning practitioners to identify and solve them in real-life time-series forecasting contexts. 
In the effort of developing the best performing machine learning model and minimizing the error across all data points, some important artifacts in the data might be overlooked. 
Moreover, data sampling methods or ensemble model approaches\cite{branco2019wercs}\cite{brancoSurveyPredictiveModeling2016} 
up until now focus on minimizing the prediction error in the underrepresented data samples and assume that the remaining data is negligible. That assumption is inaccurate, as in some applications, for example in stock market prediction, the cost of larger forecasting error in the more frequent cases could offset in the long run the potential benefit of a smaller error in a rare case.
In order to tackle real-world applications, there is a need for a broader, balanced, flexible and iterative approach, honed through interaction with domain experts and integrating the latest research in predictive models.

In this paper we propose such an approach that has been designed based on a case study in a large industrial company, targeted to forecast the temperature of a core component in a large production line. The approach involves three steps: first selecting a weight function which quantifies the sample importance; then applying one or more sampling methods to the data; and finally training and evaluating the model with and without sampling. 
In the last step, we also analyze the input importance learnt by the model using SHAP\cite{SHAP2017} to gain insights about the effect of the imbalance.
To exemplify our approach, we show how it is used for the aforementioned industrial task. We study the impact of choices in each of the steps, comparing different sampling techniques and deep learning models. In the end, we also combine the sampling with attention mechanisms \cite{Vaswani2017} to extract insights of what is learned by a deep learning model.

\section{Related Work}

The advancements in data availability from a plethora of sources, the increasing computational capacity and the progress in artificial intelligence research has led to the usage of machine and deep learning models across a multitude of applications of time-series forecasting. 
Previous work \cite{torres2021dlsurvey} surveys several use-cases including wind power forecasting, stock price prediction and estimation of remaining useful life of motors.
Despite this widespread use of models, the majority of works present specific architectures for specific datasets, while works focusing on integrated frameworks in the sense of structured approaches to a more generalized problem are more scarce. 

Although there has been an extensive amount of work in handling imbalanced datasets in classification tasks \cite{brancoSurveyPredictiveModeling2016}, regression with imbalanced data in the area of machine and deep learning has not been largely covered. In \cite{branco2019wercs}, Branco et al. study the effect of three proposed sampling methods on the predictive performance of machine learning models. In an applied example, in \cite{snieder2020}, the objective is that high water flows are predicted in a timely and accurate manner, and the problem is addressed with various sampling and ensemble methods, with an artificial neural network as a base. However, there is still a need for research into a structured approach towards real-world imbalanced regression problems, especially in the context of state of the art deep learning models.

\section{Methodology}
It is common in time-series forecasting that certain data samples have more importance than others, but are also underrepresented in the dataset, resulting in an imbalanced regression problem.
In this section we present a new approach for identifying and tackling this discrepancy with respect to imbalances in the context of regression tasks.
This approach uses a weight function that quantifies the importance of each sample which is combined with under-sampling methods to create a more balanced dataset.
The new sampled dataset is then evaluated first visually, by making density plots of the data and then numerically, by using it for training and testing a predictive model.

\subsection{Steps for Identifying and Treating Imbalanced Regression}
\label{sec_framework}
\begin{figure}
\centering
\includegraphics[width=\linewidth]{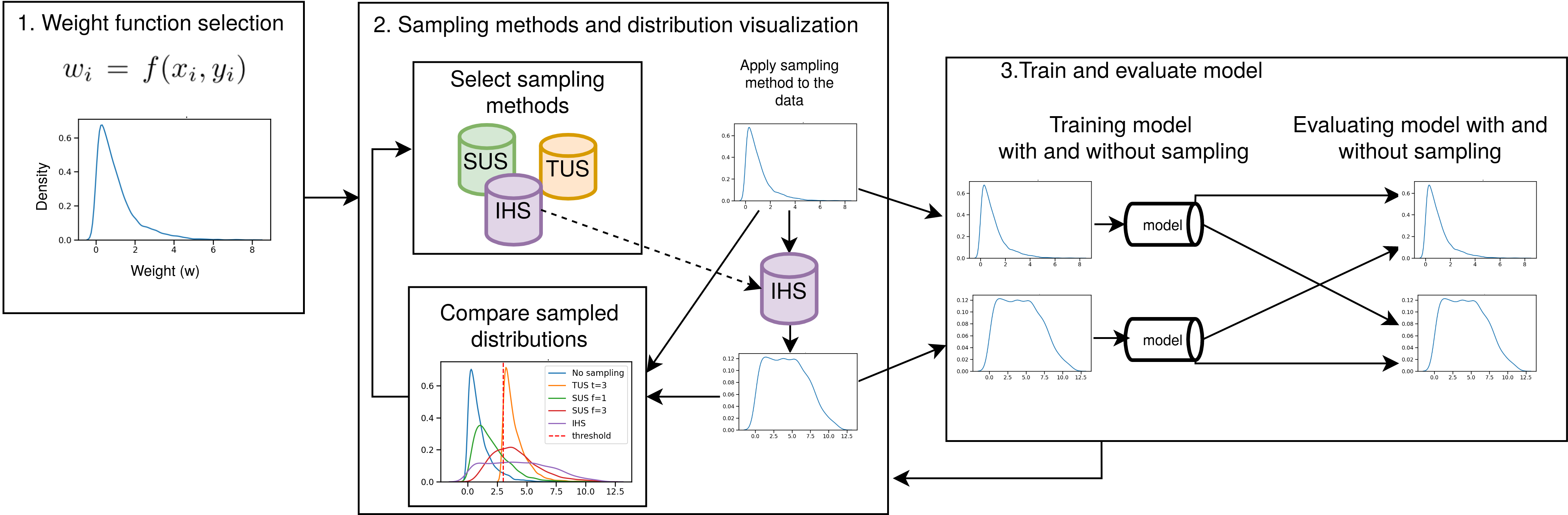}
\caption{Flowchart of the framework showing the three steps and how they interact with each other.}
\label{fig_flowchart}
\end{figure}

We propose a set of general steps for approaching the imbalance on time-series forecasting problems, which has been defined based on experiences we have collected in applying machine learning in a large scale industrial company. 
It consists of three steps illustrated in Figure \ref{fig_flowchart}.
The first step is to select or define a weight function $w_i$ to quantify the sample importance, which allows us to identify and compare the different regions of interest in the data.
The second step consists of selecting one or more sampling methods based on the weight function, applying them to the data and comparing the resulting distribution of $w_i$ against the original data using density plots.
Finally, in the third step, a predictive model is trained and evaluated using both the sampled and the original versions of the dataset.
A feedback loop can take the user from step 3 back to step 2 if the current combination of the selected sampling methods and models does not provide a satisfying performance after evaluation.
We provide more details about each step in the following sections.

\subsubsection{Step 1: Weight Function Definition}

In the example of forecasting the temperature of a motor, there can be several days where the temperature is stable, with only small fluctuations, and only a few days where the temperature increases or decreases largely. Let us assume a use-case where the user is interested in building a model to predict accurately these rarer moments when the temperature changes more than usual. In such example, the daily temperature variation can be computed as a function of the data, which we refer in our framework as the weight function. In addition, we say that the data points mapped to a high variation by the weight function belong to a region of interest.

In general, the weight function can depend either only on the target variable (or a transformation of it), only on a subset of the input variables (e.g. to signify working points of interest) or on a combination of input and output variables, to express complex regions of interest.
It can be written as $w_i = f(x_i, y_i)$, where $x_i$ and $y_i$ refer to the input (a vector for multivariate input) and the prediction target, respectively.

Equation \ref{eq_target_var} gives an example of such a function for the target variation in the context of time-series forecasting. 
The weight $w_t$ models the variation of the forecast target $y$ over the forecast horizon $\Delta$ at time step $t$. 

\begin{equation}
    \label{eq_target_var}
     w_t = \lvert y(t+\Delta) - y(t)\rvert
\end{equation}

\subsubsection{Step 2: Application of Sampling Method}
At this step, a sampling method is applied to the dataset and its effects are analyzed.
The sampling is based on the weight function previously selected and the relative proportions that the user wants to keep for the different regions of interest. 
We identify three scenarios for this step and we propose an under-sampling method for each one of them.
\begin{itemize}
    \item \textbf{Threshold under-sampling:} when the user identifies which region of the weight function is not important for the forecast task and can be removed;
    \item \textbf{Stochastic under-sampling:} when some regions are more important than others, but none can be entirely discarded;
    \item \textbf{Inverse histogram under-sampling:} when all regions are equally important and the user wants to have a balanced distribution of data over all of them.
\end{itemize}

Threshold under-sampling (TUS) consists of removing all samples that lie below a given threshold of the weight function, and all the remaining samples have the same chance of being selected. 
This method is suitable for the cases where the user knows exactly what is the region of interest is in order to be able to select the best threshold, and it assumes that the samples below the threshold aren't interesting to the prediction task.
Equation \ref{eq_thresh_prob} expresses the chance of sampling data point $t$ given its weight $w_t$.
 
\begin{align}
     \text{TUS}(w_t) &= 
    \begin{cases}
        1,  & \text{if } w_t > \tau\\
        0,  & \text{otherwise}
    \end{cases} \label{eq_thresh_prob}\\
     \text{SUS}(w_t) &=  w_t^f \label{eq_sampling_prob}\\
    \text{IHS}(w_t) &= h^{-1}(w_t)
\label{eq_hist_sampling}
 \end{align}

Stochastic under-sampling (SUS) uses the weight $w_t$ computed for each data point as the probability of sampling it.
Different from the TUS, SUS allows every sample, even the ones with lower weight value, to be sampled to avoid the creation of a new imbalance against those samples.
Equation \ref{eq_sampling_prob} models the relative probability of sampling a window from the dataset at time $t$ by using SUS.
The factor $f$ is used to increase (or decrease) the effect of the weights, thus emphasizing the more interesting moments which might be underrepresented in the data.

Inverse histogram under-sampling (IHS) is an automatic method to obtain a sample where data is approximately uniformly distributed across the selected weight function $w_t$.
It consists of building a histogram of the values of $w_t$ in the dataset and taking the inverse of the frequency of each value as the chance of sampling it.
It ensures that each $w_t$ will be under-sampled proportionally to its original frequency, so the most common values will have lower chance, while the rarer values will have higher chance.
In Eq. \ref{eq_hist_sampling} we can see a formalization of the method, where $h(w_t)$ represents the frequency of $w_t$ in the data histogram.

A good approach to gain insight about the data and the result of the sampling method is to compare the density plot of the weight function with and without using the sampling.
Such a plot can show the regions which are over or underrepresented in the data, and can also give insights about how to tune the parameters of the selected method or which method should be selected.

In some real-life cases, it might not be easy to infer directly which of the three scenarios fits the problem better.
In those cases, a subset of these methods can be selected for the next step, where we provide a heuristic to select the final method.

\subsubsection{Step 3: Predictive Model Training and Evaluation}
Finally, at this step, we can assess how much a predictive model improves by using the selected sampling methods.
For that, we train and evaluate the model with and without using the sampling method on a separate evaluation set, so we end up with different combinations of training and evaluation sets which we will use to contrast the obtained evaluation errors and determine if the model benefits from the sampling.
For the cases where the goal is to have a model that performs well on the samples of higher weight without sacrificing the performance on the rest of the data, we propose a heuristic for selecting the final sampling method to train the predictive model based on the results of the different evaluation sets.
It is defined as:
\begin{enumerate}
    \item For each sampling method, sample a training set and train a model with it;
    \item For each sampling method, sample a separate evaluation set and evaluate the trained model on it;
    \item Make a list of highest error over all the evaluation sets of each trained model to get an upper bound on its RMSE error;
    \item Select the model with the lowest error in the list.
\end{enumerate}

Next to studying the impact of the sampling on the performance of the models, we also propose to study how the models themselves change by using SHAP\cite{SHAP2017}, which is a model agnostic technique. 
SHAP gives the relative importance of each input feature to the output of the model which can be compared when the model is trained with and without the sampling.

In addition, we take advantage of deep learning models with attention mechanisms \cite{Vaswani2017} to gain extra insights of what is learnt. 
As an example of an attention-based model, TACN\cite{pantiskas2020tacn} is a deep neural network model that provides the importance of the input time-series across time steps through an attention mechanism.
The change of the patterns shown by the mechanism also provides insights into the sampling effects on the model.

\section{Experimental Setup}
\label{sec_experimental_setup}
To give a real-life example of our approach, in this section we present a case to evaluate it based on a motor temperature prediction dataset. 
We also explain the techniques used at each step of the experiments and why they were chosen.

\subsection{Motor Temperature Dataset}
The dataset used in this experiment is made of sensor measurements extracted from a steel processing conveyor belt.
The prediction target is the temperature of a bridle motor, which should be forecasted 5 minutes in advance to allow the operators to take preventive actions before a possible overheat.
The rest of the data consist of properties of the steel strip (i.e. width, thickness and yield), the speed of the line, the tension applied by the bridles, the current temperature measurements of the motor, among others.
The sensors are sampled every 10 seconds, and there are in total about 2 million samples.

In this dataset, we identify the temperature variation as a special property regarding the prediction target.
We analyze the dataset based on this property and follow the steps of our framework: selecting a weight function, then selecting the sampling methods, and visualizing the sampling result.

\subsection{Instantiation of the Framework}
Here we describe the choices made at each step of the framework for analyzing the imbalance of the temperature variation.

\subsubsection{Step 1 - Temperature Variation Weight Function}
The temperature variation is an important property to this forecasting task since the predictive model must predict accurately when the temperature will rise.
Even if, on average, the model has a satisfying performance, it may still be inaccurate when predicting higher variation if the dataset is imbalanced.
So for step 1 of the framework, we select the temperature variation as the weight function, which is modelled by equation \ref{eq_target_var}, using $\Delta$ as 30 time steps (5 minutes), which is the forecast horizon.

\subsubsection{Step 2 - Sampling Method Choice}
For step 2, we experiment with three sampling methods: SUS with factor 1, SUS with factor 3 and IHS. 
Each one under-samples a different amount of low temperature variation data, creating a different balance, as shown by Figure \ref{fig_var_dist}.
SUS with factor 1 and 3 are chosen to compare the effect of the factor in the proportion of data samples with low and high variation.
For the IHS method, we use the Freedman Diaconis estimator\cite{freedman1981histogram}
to compute the bin width of the histogram.
10.000 training data samples are extracted using each method.

\subsubsection{Step 3 - Predictive Model Choices}
In our experiments, we choose 
a multilayer preceptron\cite{rosenblatt1958perceptron} as a deep neural network baseline which has been used in time-series forecasting\cite{ahmed2010empirical} and three deep neural networks specialized in temporal data.
These specialized architectures are the long short-term memory (LSTM)\cite{lstm1997}, a popular recurrent neural network, the temporal convolutional network (TCN)\cite{baiEmpiricalEvaluationGeneric2018}, a sequence-to-sequence model which has shown promise when trained on a large amount of data\cite{silvestrin2019comparative} and the temporal attention convolutional network (TACN)\cite{pantiskas2020tacn}.

The TACN is an architecture which combines a TCN with an attention mechanism\cite{Vaswani2017} to achieve interpretable and accurate forecasting.
The per-instance interpretability comes in the form of a vector, equal to the input window size, which shows the importance of each input step to the forecasting output. The higher the value of the vector at a specific step, the higher the contribution of the input value at that step to the final output. By scaling the vector to the 0-1 range, we can estimate the relative importance among the input steps. Although this vector is produced per instance, we can draw conclusions about the generic learned behavior of the model by collecting and analyzing the vectors from a large number of instances.

For data pre-processing, we extract a window of 5 minutes (or 30 time steps) for each sample, which is the input for the TCN, LSTM, and TACN models.
For the MLP model, we extract basic features of each sensor such as the mean, standard deviation, minimal and maximal values for each window.
We also keep the last time step as an additional feature and for later analysis of the temperature variation case.
All the models are evaluated using the root-mean-square error metric (RMSE).

\section{Results}
In this section we describe the results obtained after applying our framework starting from step 2. 
Step 1 is already defined in Section \ref{sec_experimental_setup}.

\subsection{Step 2 - Comparison of the Sampling Methods}
Figure \ref{fig_var_dist} shows the variation distribution after applying the sampling methods.
Without any sampling, the dataset has a strong bias towards samples with variation close to zero, meaning that the temperature is stable, or varies very slowly most of the time.
SUS with factor 3 give more emphasis to samples with higher variation, while significantly reducing the number of samples with lower variation.
The sampling using SUS with factor 1, on the other hand, is more conservative and preserves a considerable amount of samples with low variation.
Finally, IHS gives the best balance across all values, and is the one which gives the highest proportion of samples in the extreme of the temperature variation spectrum (above 6 degrees in Figure \ref{fig_var_dist}).

\subsection{Step 3 - Analysis of the Results}
The results of the four models trained and tested with the selected sampling methods based on temperature variation can be seen in Table \ref{tab:var-results}, with the lower error per evaluation set highlighted.
The effect of the imbalance of the original data distribution is clearly shown in the "None" rows, where the models were trained without sampling.
For those lines, the RMSE is much higher in the SUS 3 column, where there is a smaller number of samples of low variation, suggesting that the models are biased towards low variation samples if trained without sampling methods.

\begin{figure}
\subfloat[][]{
\centering
\includegraphics[width=0.3\linewidth]{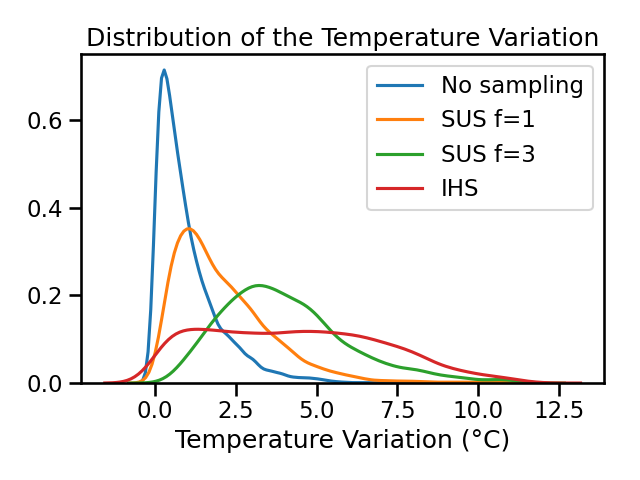}
\label{fig_var_dist}
}
\hfill
\subfloat[][]{
\centering
\includegraphics[width=0.3\linewidth]{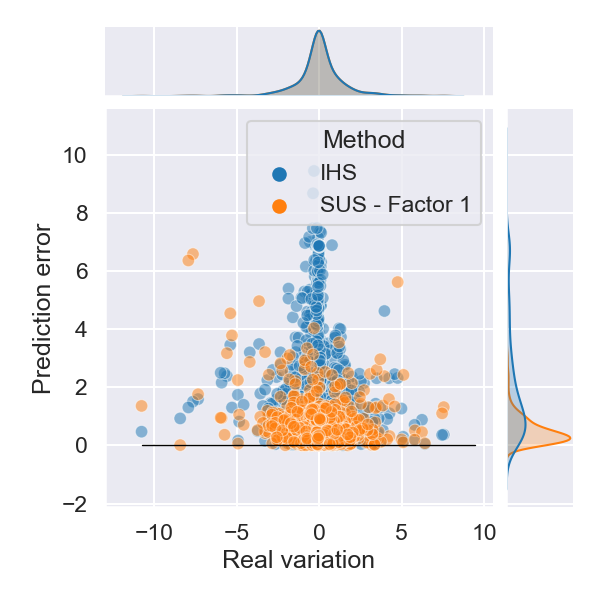}
\label{fig_rand_var_scatter}
}
\hfill
\centering
\subfloat[][]{
\centering
\includegraphics[width=0.3\linewidth]{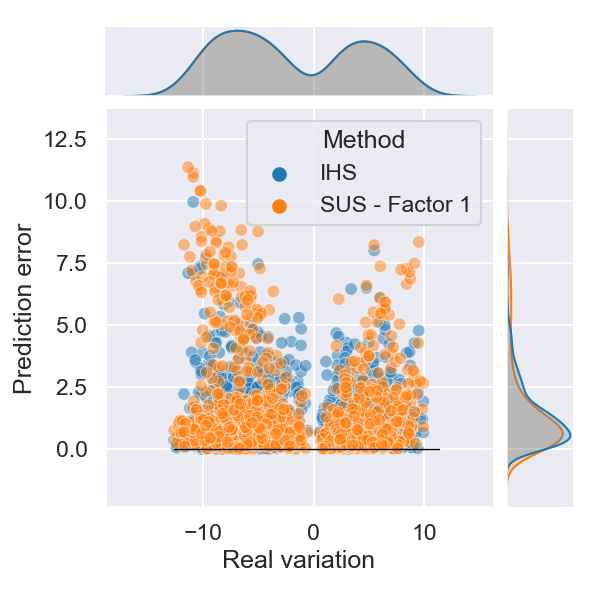}
\label{fig_high_var_scatter}
}
\caption{(a) Comparison of different sampling methods using the temperature variation as weight function. (b) Prediction error for TCN model trained using IHS and SUS with factor 1 on "no sampling" evaluation set and (c) on SUS with factor 3 evaluation set.}
\end{figure}

On the other hand, these results show that there is a trade-off between favoring samples with and without temperature variation.
Models trained with a more aggressive kind of sampling, such as SUS with factor 3, have a much higher error when evaluated on the unsampled data
than the models trained with SUS factor 1, for example.
This can be explained by the density difference between samples with low variation (below 2.5 degrees in Figure \ref{fig_var_dist}), the same samples that are more common in the "no sampling" dataset. 
With our approach, this trade-off which exhibits non-linear behavior can be estimated, taking into account the end-user preferences, and it can lead to a re-evaluation of the sampling method in step 2. Also, together with these metrics, using insights about the model as described later in this subsection can indicate the sampling method that leads to the most encompassing, generalizable patterns learned by the models, thus creating a balance for the performance across data samples. 

\begin{table}[ht]
\centering
\caption{RMSE results for different sampling methods based on temperature variation. "None" means that the model was trained or evaluated without using a sampling method.}
\label{tab:var-results}
\begin{tabular}{|c|c|c|c|c|c|}
\hline
Model               & Trained on              & \multicolumn{4}{c|}{Evaluated on}    \\ \hline
                    &                         & \textbf{None}    & \textbf{SUS - 1} & \textbf{SUS - 3} & \textbf{IHS}            \\ \hline
MLP                 & \textbf{None}    & 1.401 $\pm$ 0.181 & 2.270 $\pm$ 0.177          & 3.611 $\pm$ 0.322         &  2.984 $\pm$ 0.241       \\ \hline
                    & \textbf{SUS - 1} & 1.704 $\pm$ 0.289          & 2.085 $\pm$ 0.239 & 2.886 $\pm$ 0.269          & 2.495 $\pm$ 0.234        \\ \hline
                    & \textbf{SUS - 3} & 4.150 $\pm$ 0.635          & 3.089 $\pm$ 0.390          & 2.430 $\pm$ 0.305  & 2.836 $\pm$ 0.32          \\ \hline
                    & \textbf{IHS}            & 3.066 $\pm$ 0.401          & 2.728 $\pm$ 0.228          & 2.539 $\pm$ 0.248          & 2.663 $\pm$ 0.217          \\ \hline
LSTM                & \textbf{None}    & 1.032 $\pm$ 0.091 & 1.857 $\pm$ 0.112          & 3.275 $\pm$ 0.129          & 3.275 $\pm$ 0.13        \\ \hline
                    & \textbf{SUS - 1} & 1.283 $\pm$ 0.153          & 1.469 $\pm$ 0.123 & 2.769 $\pm$ 0.130          & 2.731 $\pm$ 0.094        \\ \hline
                    & \textbf{SUS - 3} & 3.595 $\pm$ 0.268          & 2.549 $\pm$ 0.128           & \textbf{1.464 $\pm$ 0.24}  & 2.415 $\pm$ 0.095          \\ \hline                    
                    & \textbf{IHS}            & 2.728 $\pm$ 0.174          & 2.131 $\pm$ 0.093          & 1.863 $\pm$ 0.106          & 2.27 $\pm$ 0.093           \\ \hline
TCN                 & \textbf{None}    & \textbf{0.871 $\pm$ 0.021} & 1.684 $\pm$ 0.037          & 3.060 $\pm$ 0.068    & 3.142 $\pm$ 0.05          \\ \hline
                    & \textbf{SUS - 1} & 1.007 $\pm$ 0.079          & \textbf{1.462 $\pm$ 0.041} & 2.686 $\pm$ 0.066          & 2.703 $\pm$ 0.063          \\ \hline
                    & \textbf{SUS - 3} & 3.41 $\pm$ 0.213           & 2.4 $\pm$ 0.091            & 1.592 $\pm$ 0.124 & 2.283 $\pm$ 0.008          \\ \hline                
                    & \textbf{IHS}            & 2.579 $\pm$ 0.231          & 2.016 $\pm$ 0.091          & 1.845 $\pm$ 0.062          & \textbf{2.145 $\pm$ 0.039}           \\ \hline
TACN                & \textbf{None}    & 1.171 $\pm$ 0.016        & 2.637 $\pm$ 0.051          & 5.167 $\pm$ 0.1            & 5.064 $\pm$ 0.101          \\ \hline
                    & \textbf{SUS - 1} & 1.334 $\pm$ 0.242          & 2.077 $\pm$ 0.355          & 3.878 $\pm$ 0.675          & 3.183 $\pm$ 0.694          \\ \hline
                    & \textbf{SUS - 3} & 3.802 $\pm$ 0.538          & 2.786 $\pm$ 0.428          & 2.36 $\pm$ 0.758           & 2.883 $\pm$ 0.547          \\ \hline
                    & \textbf{IHS}            & 3.093 $\pm$ 0.918          & 2.424 $\pm$ 0.608          & 2.430 $\pm$ 0.748          & 2.752 $\pm$ 0.623           \\ \hline
                    
\end{tabular}
\end{table}

\begin{table}[ht]
\centering
\caption{The maximum error obtained by the TCN model over all the evaluation sets. Each line corresponds to a training set obtained from a different sampling method.}
\label{tab:tcn-results}
\begin{tabular}{|c|c|c|}
\hline
Trained on              & Max. error  & Measured on       \\ \hline
\textbf{No sampling}    & 3.142 $\pm$ 0.05    & IHS          \\ \hline
\textbf{SUS - 1} & 2.703 $\pm$ 0.063 & IHS           \\ \hline
\textbf{SUS - 3} & 3.41 $\pm$ 0.213  & No sampling           \\ \hline                    
\textbf{IHS}          & 2.579 $\pm$ 0.231            & No sampling           \\ \hline
                    
\end{tabular}
\end{table}

Since the results show that the TCN achieves a relatively lower error in all the evaluation sets, we select it as the best model and follow the heuristic described in section \ref{sec_framework}.
Table \ref{tab:tcn-results} shows the maximum error obtained by it over all the evaluation samples.
The two lowest RMSE values reported in that table are from the TCN model trained with SUS factor 1 and IHS, and the evaluation sets where they have the highest error are \emph{No sampling} and \emph{IHS}.
By comparing the performance of the TCN trained with both methods on the evaluation set without sampling, we can clearly see that the model trained with SUS with factor 1 has lower spreading of the error (Fig. \ref{fig_rand_var_scatter}).
On the other hand, the same comparison on the evaluation set with higher variation (Fig. \ref{fig_high_var_scatter}) shows that both models have similar error spreading, and the small advantage of using IHS in this case does not compensate for the increase in error in the low variation samples.
Therefore, we can conclude that SUS with factor 1 is the sampling method with best performance across samples with low and high temperature variation.

\subsubsection{Imbalance Effects in DL Models}
\label{sec_model_analysis}
In the third step of the framework, we also verify how the imbalances 
affect the performance and learned patterns of the predictive models,
when they are trained on data with sampling versus unsampled data.
To do that, we focus on the temperature variation property of the dataset, and we measure the SHAP values of the MLP model, as well as the attention importance values of the TACN.

\begin{figure}[ht]

\includegraphics[width=\linewidth]{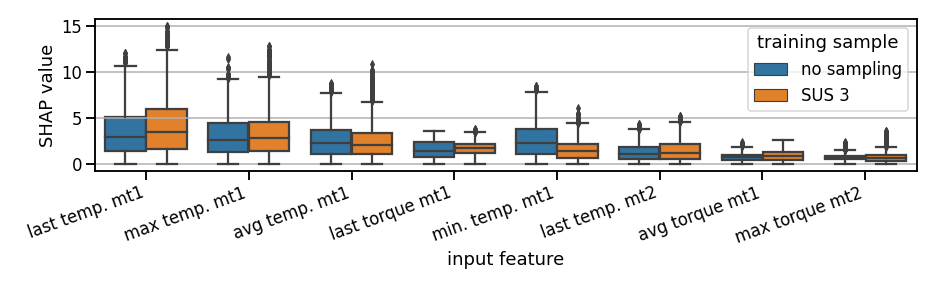}
\caption{Comparison of the absolute SHAP values estimated for the MLP model trained with SUS factor 3 and without sampling. The values are computed based on inputs from the evaluation set sampled using SUS factor 3. The input features are computed over a window of 5 minutes. \emph{mt1} and \emph{mt2} correspond to the first and second motors of the bridle set.}\label{fig_mlp_shap}
\end{figure}

To assess how the sampling methods influence the MLP model,
we extract its SHAP values using the SUS 3 evaluation set.
We compare both the MLP trained with SUS 3 and without sampling.
Figure \ref{fig_mlp_shap} shows that both models rely mostly on the last temperature measurement to make the forecast.
This could be explained by the fact that the last temperature is relatively close to the predicted temperature, even when there is high variation.
One hypothesis for such fact is that the MLP does not handle the time dependency of the inputs and, thus has a disadvantage in comparison to other temporal models such as the TCN or the LSTM.

To gain insights about the differences in the behavior of the trained TACN models using the interpretability mechanism, we run inference on the SUS with factor 3 evaluation set for the models trained on (a) unsampled data and (b) on the SUS factor 3 train set, and we study the resulting attention pattern variance.

\begin{figure}[ht]
\begin{center}
\includegraphics[width=0.8\linewidth]{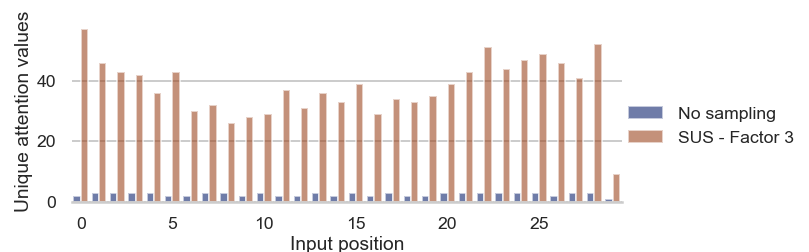}
\caption{Unique attention values per input step from sample TACN models trained on data with (a) no sampling and (b) SUS with factor 3}
\label{fig_bar_plot_att_values}
\end{center}
\end{figure}

In order to quantify this variance, we enumerate for both models the unique learned attention values for each input time step across all test samples, rounded to the second decimal, and present the result on Fig. \ref{fig_bar_plot_att_values}. For the model trained on unsampled data, the unique values for each position are at most 3, while for the SUS model they are between 30 and 50. The above observations lead us to the following conclusions: The model trained on the unsampled data has learned a high reliance on the last value and limited number of patterns, which serves well in minimizing the error for the majority of the samples but results in low performance on the large variation samples. In contrast, the model trained on the SUS data is forced to learn a larger variety of patterns to accommodate for this target variation. 

\section{Conclusion}
We presented a framework to analyze imbalanced time-series forecasting problems and to train and evaluate ML models taking into account important properties.
To our knowledge, this is the first framework which provides clear steps to help practitioners to select and compare different sampling methods and predictive models for such problems.
It is put into practice to forecast the temperature of a motor in a steel processing conveyor belt, based on data extracted from a real-world industrial process and is validated in cooperation with domain experts. 
The problem analysis is made through the lens of the temperature variation property. 
We study the dataset using three different sampling methods and we train four different DL models to evaluate and compare the effectiveness of each combination of sampling and model.
We also show the imbalance of the temperature variation and how it changes the models' predictions when they are trained with different proportions of samples with high temperature variation.
Finally, we use SHAP values and the TACN model's attention mechanism to show the effect of low temperature variation in the dataset on the forecast models, inducing them to rely mostly on the last observed temperature.

As future work, our framework could be put into practice to analyze new time-series prediction tasks, combining with more sampling techniques.
The framework can be extended with new weight functions, making it suitable for an even wider range of tasks.
In addition, our results point out to a possible relationship between the prediction error and the distribution of the training data which might be worth investigating.

\section*{Acknowledgements}
This work has been conducted as part of the Just in Time Maintenance project funded by the European Fund for Regional Development.
We also thank Tata Steel Europe for providing the data and technical expertise required for our experiments.

%
%
%
\bibliographystyle{splncs04}
\bibliography{references}
\end{document}